\documentclass[final]{cvpr}

\usepackage{times}
\usepackage{epsfig}
\usepackage{graphicx}
\usepackage{amsmath}
\usepackage{amssymb}
\usepackage{xcolor}
\usepackage[switch]{lineno}  
\usepackage{amsfonts,amsmath,amssymb}       
\usepackage{adjustbox}
\usepackage{multirow}
\usepackage{booktabs}
\usepackage{soul,colortbl}
\usepackage{diagbox}
\usepackage{capt-of}
\usepackage{pifont}
\graphicspath{{figures/}}

\definecolor{tblue}{HTML}{174992}
\definecolor{Gray}{gray}{0.85}

\newcommand{\RN}[1]{%
	\textup{\lowercase\expandafter{\it \romannumeral#1}}%
}
\ifx\remark\undefined
\newtheorem{remark}{Remark}
\fi
\newcommand{\cmark}{\ding{51}}%
\newcommand{\xmark}{\ding{55}}%

\usepackage[pagebackref=true,breaklinks=true,colorlinks,bookmarks=false]{hyperref}

\def\cvprPaperID{6353} 
\def\confYear{CVPR 2021}

\begin{document}

\title{ReMP: Rectified Metric Propagation for Few-Shot Learning}

\author{Yang Zhao$^1$ \quad Chunyuan Li$^2$ \quad Ping Yu$^1$ \quad Changyou Chen$^1$\\
$^1$University at Buffalo \quad $^2$Microsoft Research \\
}

\maketitle
\begin{abstract}
Few-shot learning features the capability of generalizing from a few examples. In this paper, we first identify that a discriminative feature space, namely a rectified metric space, that is learned to maintain the metric consistency from training to testing, is an essential component to the success of metric-based few-shot learning. Numerous analyses indicate that a simple modification of the objective can yield substantial performance gains. The resulting approach, called \textit{rectified metric propagation} (ReMP), further optimizes an attentive prototype propagation network, and applies a repulsive force to make confident predictions. Extensive experiments demonstrate that the proposed ReMP is effective and efficient, and outperforms the state of the arts on various standard few-shot learning datasets.
\end{abstract}

\section{Introduction}
Learning and generalizing from a few examples to effectively make predictions in new domains/tasks is a common scenario in real applications, as the supervised information is often hard to acquire due to some practical considerations, such as high labeling cost, privacy, safety or ethic issues.
Naively applying the traditional supervised learning techniques tend to overfit in this scenario~\cite{finn2017model,vinyals2016matching}. This stimulates the emergence of {\em few-shot learning (FSL)}, which mimics the human ability of recognizing new data after observing a few instances.

In such a few-shot regime, it is desired to leverage knowledge (\textit{e.g.} parameters and embeddings) from one task to another. Recent research has shown promising results in exploiting \textit{episodic learning} on this task. In essence, the episodic learning aims to mimic the real testing scenarios where the model generalizes information from a few labeled instances 
(called \textit{support set}) to predict labels of unlabeled instances (called \textit{query set}) in each episode. Research efforts in solving FSL can be broadly categorized into two types:
$(\RN{1})$ \textit{Meta-learning}~\cite{finn2017model,nichol2018reptile} can quickly adapt parameters to a new task after fine-tuning; 
$(\RN{2})$ \textit{Embedding and metric-based learning}~\cite{sung2018learning,vinyals2016matching} directly performs nearest-neighbor classification given a similarity metric on an embedding space. Our work is falling into this category.

The fundamental task for the metric learning methods is to design an appropriate metric space~\cite{oreshkin2018tadam} that can satisfy the following three properties simultaneously: $(\RN{1})$ a valid measure (or metric) such as Euclidean distance and Cosine similarity~\cite{snell2017prototypical,oreshkin2018tadam,kye2020transductive}); $(\RN{2})$ a cost function that can fully exploit input information to update the network; $(\RN{3})$ the metric consistency from training to testing, as well as from pretraining to fine-tuning if a pretraining is being applied. A contradictory example that arouses an inconsistency issue is seen in most works as well as recent advanced metric-based FSL frameworks~\cite{chen2020new,ye2018fewshot}. They adopt an inefficient two-phase training procedure: a \emph{parametric} pre-training with a linear classifier and a \emph{non-parametric} fine-tuning with the nearest-neighbor prediction. In FSL, the pretraining essentially leads to a different metric space from that learnt by fine-tuning. However, the testing is only accessible to use the \emph{non-parametric} nearest neighbor classifier (see more details in the Background section). In the end, the inconsistency is reflected on the very limited performance gain~\cite{ye2018fewshot}.

While the first two properties have largely been individually explored~\cite{snell2017prototypical,qi2018low,lifchitz2019dense}, the last one is less studied (if not ignored at all).
We observe that this missing ingredient can lead up to 5\% absolute recognition accuracy drop when the inconsistent metric exists. In the meantime, however, this kind of parametric linear classifier is essential to explore a discriminative metric space and to encourage faster convergence. So the natural question is: \emph{is there a strategy that can maintain the metric consistency as well as fully utilize the training data information to learn a meaningful metric space?}

To this end, we argue that a suitable interaction of above three properties can yield significant improvements in terms of performance and stability. We study the FSL problem from a unified perspective and propose a rectified metric propagation (ReMP) framework that can progressively make predictions in a discriminative feature space, namely the rectified metric space. The rectified metric is induced from maintaining consistency from training to testing. Specifically, we firstly design a \textbf{cooperative learning} objective that considers training-testing consistency both within a global parametric classifier (called \textbf{global matching}) and local nearest neighbor prediction among instances (called \textbf{local matching}). The framework proceeds in a feed-forward pass without either pre-training or fine-tuning. To further enhance the nearest neighbor prediction confidence, we propose an attention-based contextualized label embedding method to iteratively rectify the prototypes, such that data importance is taken into account when calculating class prototypes. The resulting approach is simpler and more efficient than related recent approaches. 
The contributions of this paper are threefold:
\begin{itemize}
    \item We identify the metric inconsistency issue between training and testing in FSL, a long-standing issue of almost all metric-based FSL approaches. To alleviate this problem, we propose a \textbf{cooperative label-aligned training} scheme, where the unabridged metric space can be inherited to the testing phase. 
    \item We describe a \textbf{contextualized label embedding module containing attentive prototype propagation layers} to take into account data importance to make more confident predictions.
    \item With the above two novelties, \textbf{new state-of-the-art (SoTA)} performances on four standard FSL classification datasets, the \textit{miniImageNet}, \textit{tieredImageNet} and \textit{CIFAR-FS} are achieved.
\end{itemize}
The rest of the paper is organized as follows. We first summarize the related work and how our proposal
is differentiated from those. Then, we describe our contribution in detail and present extensive experiments to demonstrate our justifications. Finally, we conclude our paper and highlight future research directions. 

\section{Related work}
In this section, we introduce related works and make distinction between our approach and related FSL methods. These methods broadly fall into three categories, meta-learning, embedding and metric learning, and transductive learning based approaches.

\paragraph{Meta-learning}
Meta-learning~\cite{thrun1998lifelong,finn2017model}, or learning-to-learn, is a framework that is capable of learning a task-specific meta-network. After observing the support set of a new task, the meta-network can quickly adapt to be evaluated on the query set of that new task. \cite{ravi2016optimization} proposes to finetune an LSTM-based optimizer besides the meta-learner to maximize the performance. \cite{munkhdalai2017meta} learns to change its inductive bias via fast parameterization. These works include MAML~\cite{finn2017model}, Reptile~\cite{nichol2018reptile}, Meta-SGD~\cite{li2017meta}, Bayesian-MAML~\cite{yoon2018bayesian}, Implicit-MAML~\cite{rajeswaran2019meta} and LEO~\cite{rusu2018meta}. However, the aforementioned approaches often suffer from over-fitting and sensitive to architectures, making the performance after fine-tuning on a new task limited. Although some recent work \cite{antoniou2018train} proposes some modifications, this line of approaches is still unsatisfactory for solving FSL. By contrast, our proposed model can make predictions in a direct and efficient feed-forward manner without the necessity of fine-tuning. 
Recently, perhaps the most popular approach is model agnostic meta learning (MAML) \cite{finn2017model}. MAML learns a meta-parameter-initialization of a meta-network such that it can solve a new task with only a few gradient descent steps.

\paragraph{Embedding and metric learning}
This class of methods for FSL has drawn more and more attention.  The main goal is to learn a transferable and discriminative feature space that preserves the neighborhood structure, {\it e.g.}, the Matching network \cite{vinyals2016matching} and Prototypical network \cite{snell2017prototypical}. This means objects belonging to the same class should be consistently closer to each other in the feature space measured by some similarity measures, {\it e.g.}, the Euclidean distance and Cosine similarity, and vice versa. Two typical works Relational network \cite{sung2018learning} and TADAM~\cite{oreshkin2018tadam} made further improvements. Another line of work proposes to learn global prototypes \cite{gidaris2018dynamic,qi2018low}.
Label embedding has been proposed as anchor points to improve text classification in NLP~\cite{wang2018joint}. 
Different from these approaches, the proposed cooperative learning objective can fully \emph{exploit} the manifold information of both support set and query set in training, and progressively \emph{explore} a discriminative feature space to refine the prototype in testing.

\paragraph{Transductive learning}
The transductive learning is first introduced in TPN \cite{liu2018learning} that exploits the manifold structure in the data by learning a graph construction to propagate labels from the support set to the unlabeled query set. It alleviates the small-data problem in FSL and has been shown to outperform the inductive learning (inaccessible to a query set) counterparts. \cite{kim2019edge} further enhances the graph construction module. FEAT~\cite{ye2018fewshot}, CAN~\cite{hou2019cross}, Meta-Fun~\cite{xu2019metafun} and EPNet~\cite{rodriguez2020embedding} optimize the embedding manifold to better generalize to unseen classes. DFMN-MCT~\cite{kye2020transductive} chooses to propagate prototypes along with assigning confidence scores to all unlabeled queries, leading to SoTA results. However, the pretraining phase of FEAT and the pixel-wise dense classifier of DFMN-MCT lead to an inconsistent pipeline from training to testing. Our work is distinguished from existing works by that we propose to rectify the metric space, which aims to close the gap between training and testing. Furthermore, our approach can achieve new SoTA results with a newly proposed repulsive attention strategy by stacking up more attention layers.

\section{Background}
\subsection{Problem definition}
Similar to the supervised-learning setting, a dataset is typically divided into three parts in FSL: a training dataset $\mathcal{D}^{\text{train}}$, a testing dataset $\mathcal{D}^{\text{test}}$ and a validation dataset $\mathcal{D}^{\text{val}}$. The main distinction of FSL is that the three sets have disjoint label spaces.  The episodic classification \cite{finn2017model,mishra2017simple,lee2019meta,vinyals2016matching,snell2017prototypical,santoro2016meta} is a common and effective approach to FSL, where the training dataset is exploited to simulate the few-shot learning setting via episode-based training.

Specifically, to characterize generalization, in the training process, one typically randomly samples $N$ classes instances in each episode, containing a \textit{support} set $\mathcal{S}=\{(\mathbf{x}_i, y_i \}_{i=1}^{N\times K}$ ($K$ samples per class) and a \textit{query} set $\mathcal{Q}=\{(\mathbf{\tilde{x}}_i, \tilde{y}_i \}_{i=1}^{N\times M}$ ($M$ samples per class). {\em The support set is used to calculate prototypes and then make predictions on the query set, which in turn is used to update the model.} This setting is often abbreviated as {\em {\it N}-way {\it K}-shot} FSL.
The number of $K$ is generally very small, \textit{e.g.} 1 and 5. The goal of FSL is to learn a model to exploit such a low resource data 
set $\mathcal{S}$ to predict labels for queries in $\mathcal{Q}$.

\subsection{Revisiting the classifiers in the feature space}
Current approaches to FSL resort to a two-step procedure, represented as a function decomposition $f \circ g $, where
$(\RN{1})$ $f: \mathbf{x}\xrightarrow[]{} \mathbf{z}$ is a feature extractor, which maps an input $\mathbf{x} \in \mathbb{R}^D$ to a feature vector $\mathbf{z} \in \mathbb{R}^d$ in a transferable and discriminative space. It is worth noting that $d$ is not necessarily much smaller than $D$ and it is mostly related to the complexity of $f$; 
$(\RN{2})$
$g$ is a classifier, mapping the feature $\mathbf{z}$ into a vector of logits $g(\mathbf{z})$ and produce a probability distribution $p(\cdot|\mathbf{x})$ over all categories activated by the \textsf{Softmax}. 
Since $f$ and $g$ are sometimes not trained end-to-end, it is crucial to find a good association between a classifier $ g$ and the feature representation $\mathbf{z}$. Typically, there are two general paradigms to design the classifier $g$. Unfortunately, both paradigms endow a limitation. Our proposed approach would inherit the best of both worlds, as described in the next section.
\begin{itemize}
    \item {\bf Parametric Methods.} These approaches usually learn a linear classifier $g(\mathbf{z})= \mathbf{w}^\intercal \mathbf{z} + \mathbf{b}, \mathbf{w}\in \mathbb{R}^{d \times N}, \mathbf{b} \in \mathbb{R}^{N}$, thus inherits fast convergence. However, since in FSL, the label space in the testing is disjoint with that in the training phase. This implies that $\{\mathbf{w, b}\}$ trained on the training data is not ready to be adopted for testing directly. Two solutions can be designed to mitigate this problem: $(\RN{1})$ Fine-tuning to adapt the parameters \cite{finn2017model,li2017meta,rusu2018meta} to test sets; $(\RN{2})$ Performing the nearest neighbor search in an embedding space \cite{qi2018low,kye2020transductive,chen2020new} for testing. We only consider the second approach in the following as the fine-tuning approaches are often unsatisfactory and time-consuming. However, the second solution would lead to \textbf{inconsistency between training and testing}. This is because $g(\mathbf{z})$ is computed based on a linear mapping in training; whereas the nearest neighbor is used in testing. In fact, the model is unaware of an appropriate metric in training. Such inconsistency could harm model performance.
    \item {\bf Non-parametric Methods.} In training and testing, one consistently applies a metric-based similarity measure $g(\mathbf{z})=\kappa(\mathbf{z}, \mathbf{c}_n)$, where $\mathbf{c}_n$ is the class prototype (prototype), usually defined as the intra-class mean over support embeddings; $\kappa(\cdot, \cdot)$ is often defined by Cosine similarity~\cite{oreshkin2018tadam}, negative Euclidean distance~\cite{snell2017prototypical,oreshkin2018tadam,kye2020transductive} and kernel-based functions~\cite{xu2019metafun,liu2018learning}. However, it has been found that these approaches \textbf{are hard to explore and learn a discriminative metric space}\footnote{A metric space means an embedding space, where one can compute similairty given a metric. It is different from the definition in mathematics, \textit{e.g.}, the Cosine similarity, which can be negative.} \cite{snell2017prototypical,oreshkin2018tadam,sung2018learning}, thus limiting its potential to reach a good convergence point (see Tab.~\ref{tab: metric}). 
\end{itemize}
\begin{figure*}[!htbp]
    \centering
    \scalebox{0.65}{
    \includegraphics{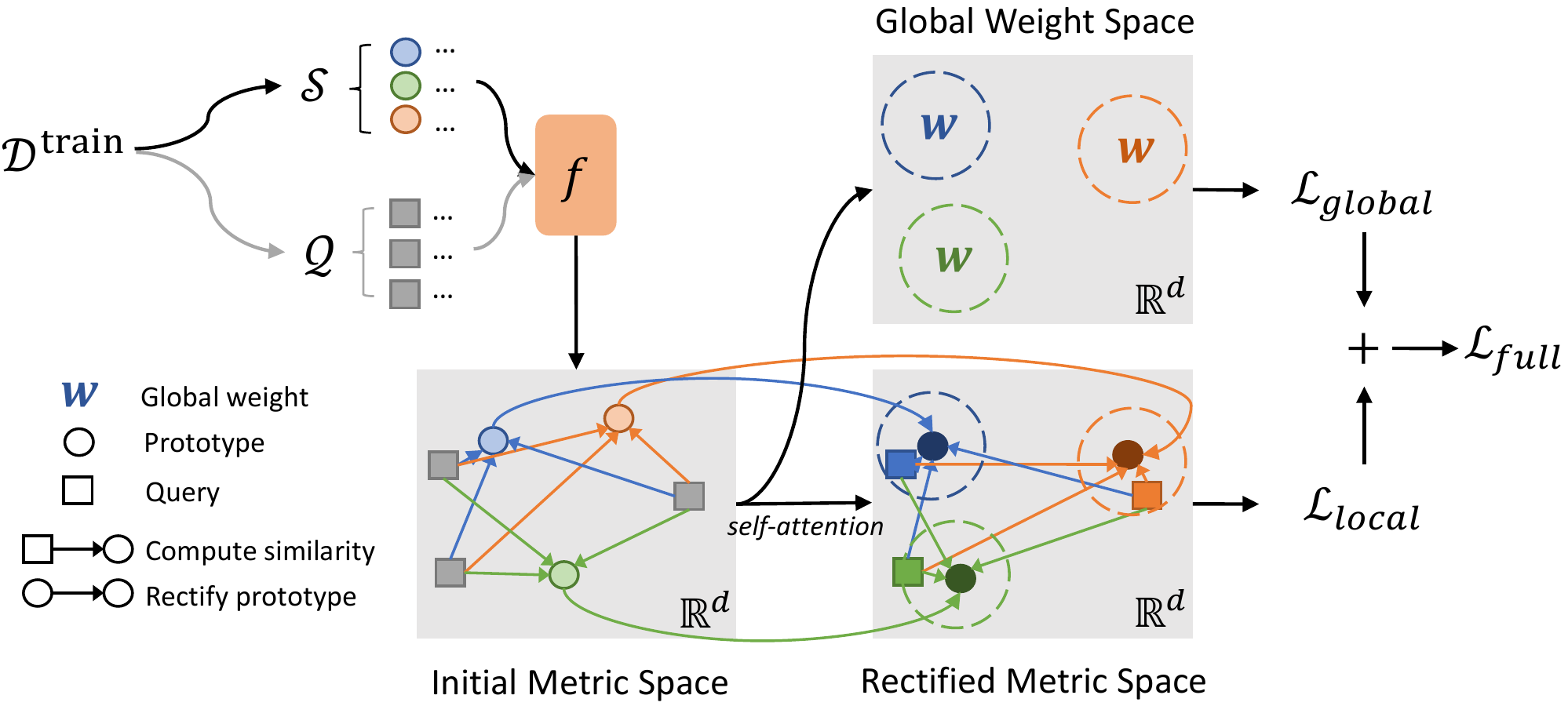}
    }
    \vspace{2mm}
    \caption{Illustration of the proposed label-aligned learning. The training considers matching in both a parametric global weight space and a non-parametric rectified metric space. In testing, predictions are only made through local matching in a feed-forward manner, where rectified prototypes help to generate a better decision boundary based on self-attention.}
    \label{fig: model}
\end{figure*}
\section{The Proposed Method}
Our goal is to overcome the two aforementioned limitations, and design a model to close the gap between training and testing, and at the same time, to learn a discriminative metric space. To this end, we first define a cooperatively label-aligned learning objective. A contextualized label embedding module with self-attention is then further formulated to iteratively propagate labels on $\mathcal{S} \cup \mathcal{Q}$, and to rectify prototypes such that more confident predictions can be made. The overall framework is illustrated in Figure~\ref{fig: model}.

\subsection{Label-Aligned Cooperative Learning}\label{sec: rectify} 
We firstly define a label-aligned learning objective to optimize the feature extractor $f$ and the classifier $g$ in training. It completes the cooperative learning on two levels, \textit{global matching} on the global prototype of $\mathcal{D}^\text{train}$ and \textit{local matching} on the local prototype of the episodic $\mathcal{S} \cup \mathcal{Q}$.

\paragraph{Notations}
To facilitate the reading, notations are clarified. The number of classes in $\mathcal{D}^\text{train}$ is $N^\text{train}$. The {\it i}-th pair of $\mathcal{Q}$ is $(\mathbf{\tilde{x}}_i, \tilde{y}_i)$.
For the $N$-way classification protocol on the instance $\mathbf{\tilde{x}}_i$, ($\tilde{y}_i^g=\tilde{y}_i, \hat{y}_i^{g}$) are the true label and predicted label ranging from 1 to $N^\text{train}$ on the global level; whereas ($\tilde{y}_i^l, \hat{y}_i^{l}$) denotes the true label and predicted label ranging from 1 to $N$ on the local level. Note $\tilde{y}_i^g$ and $\tilde{y}_i^l$ are different random variables corresponding to the same data instance $\mathbf{\tilde{x}}_i$ in a FSL setting.

\paragraph{Global matching}
To encourage fast convergence, we first define a global matching loss on $\mathcal{D}^{\text{train}}$. The global matching is a parametric model, aiming to make predictions matched with the global-level label $\tilde{y}_i^g$. To this end, we first introduce a metric $\kappa(\cdot, \cdot)$ to measure the similarity between two vectors. The metric can be flexible. We adopt the Consine similary and Euclidean distance in our experiments. The goal is to match the feature of a data instance with a set of global learnable weight vectors $\{\mathbf{w}_k \in \mathbb{R}^d\}_{k=1}^{N^\text{train}}$, each representing a global class prototype, as illustrated in Figure~\ref{fig: model}. To this end, the global likelihood on $\mathbf{\tilde{x}}_i$ is: 
\begin{align}\label{eq: ll_global}
    p(\hat{y}_i^{g}=k|\mathbf{\tilde{x}}_i) = \frac{\exp(\kappa(f(\mathbf{\tilde{x}}_i), \mathbf{w}_k))}{\sum_{k=1}^{N^\text{train}} \exp(\kappa(f(\mathbf{\tilde{x}}_i), \mathbf{w}_k))}~.
\end{align}
The global loss is then defined as the standard cross-entropy loss:
\begin{align}\label{eq: global}
    \mathcal{L}_{global}
    = -\sum_{i=1}^{NM} \sum_{k=1}^{N^{\text{train}}} \mathbb{I}(\tilde{y}_i^g=k)\log p(\hat{y}_i^{g}=k|\mathbf{\tilde{x}}_i),
\end{align}
where $\mathbb{I}(\cdot)$ is an indicator function. In each episode, all global weights are being optimized with interactive information sharing, which could significantly encourage the model's convergence. Note that, the proposed approach is different from other types of inconsistent global matching, \textit{e.g.}, the linear classifier in \cite{hou2019cross} and the dense matching in \cite{kye2020transductive}, where the embedding has a different dimension from the global weight.

\paragraph{Local matching} 
Note the above global loss $\mathcal{L}_{global}$ is \textit{parametric}, which still could cause training-testing inconsistency. To close the gap, we further incorporate a local loss on episodic $\mathcal{S} \cup \mathcal{Q}$ with $N$ classes. Instead of using a parametric model, we define a non-parametric model to directly match predictions with local prototypes. We first define the local likelihood on $\mathbf{\tilde{x}}_i$ as:
\begin{align}\label{eq: ll_local}
    p(\hat{y}_i^{l}=n|\mathbf{\tilde{x}}_i) = \frac{\exp(\kappa(f(\mathbf{\tilde{x}}_i), \mathbf{c}_n))}{\sum_{n=1}^{N} \exp(\kappa(f(\mathbf{\tilde{x}}_i), \mathbf{c}_n))}
\end{align}
where $\mathbf{c}_n$ represents the prototype for local class $n$, calculated as 
\begin{align}\label{eq: center}
    \mathbf{c}_n= \frac{1}{K}\sum_{i=1}^{NK} \mathbb{I}(y_i=n)f(\mathbf{x}_i),~~ \text{for }(\mathbf{x}_i, y_i) \in \mathcal{S}~.
\end{align}
We then define the local loss as:
\begin{align}\label{eq: local}
    \mathcal{L}_{local}
    = -\sum_{i=1}^{NM} \sum_{n=1}^{N} \mathbb{I}(\tilde{y}_i^{l}=n)\log p(\hat{y}_i^{l}=n|\mathbf{\tilde{x}}_i)~.
\end{align}
It is clear that $\mathcal{L}_{local}$ can leverage episodic $\mathcal{S}$ and $\mathcal{Q}$ for similarity comparison both in training and testing, without consulting the global parametric classifier.
\paragraph{Label-aligned objective}
To align labels in the local matching with that in the global metric space, we propose a full objective that combines Eq.~\eqref{eq: global} and Eq.~\eqref{eq: local}:
\begin{align}\label{eq: full}
    \mathcal{L}_{full} = \mathcal{L}_{global}+\alpha \mathcal{L}_{local}~,
\end{align}
where $\alpha$ is a balance hyper-parameter. At testing, predictions will only be made following Eq.~\eqref{eq: ll_local} such that the local matching is consistent for both training and testing. We claim that:
\begin{remark}\label{remark}
A good metric space is all we need in metric-based FSL. On the premise that the embedding $\mathbf{z}$ and the global weight $\mathbf{w}$ are jointly learned via Eq.~\eqref{eq: full}, global matching helps accelerate convergence and explore a space that is endowed with a strong transferability; meanwhile local matching preserves the metric inheritance from training to testing. 
\end{remark}
Besides, the metric in Equation \eqref{eq: global} and Equation \eqref{eq: local} don't need to be the same since the global weight vectors are learned rather than computed by the episodic instances. This claim is essentially different from previous methods, {\it e.g.}, CAN~\cite{hou2019cross} and MCT-DFMN~\cite{kye2020transductive}. We will give detailed justifications in the experiments.

\subsection{Contextualized Label Embeddings for Prototype Rectification}
\begin{figure}[!htbp]
    \centering
    \includegraphics[width=1.0\linewidth]{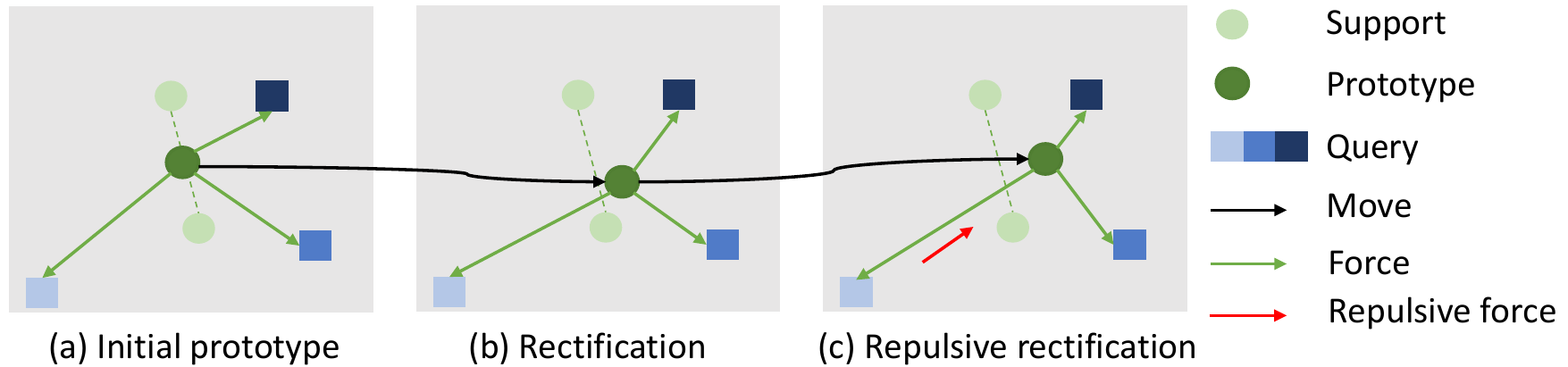}
    \vspace{2mm}
    \caption{Prototype rectification. The color depth of a query represents the attention score to the prototype. (a) The initial prototype is the mean embedding of the two instances in the support set. (b) Rectification will take the neighbor information into consideration to refine the prototype. (c) Repulsive rectification further imposes a repulsive force to get a more representative prototype.}
    \label{fig: rectification}
\end{figure}

In FSL, the training set $\mathcal{D}^{\text{train}}$ and test set $\mathcal{D}^{\text{test}}$ have disjoint label space. This makes the resulting embedding space trained on the training dataset not discriminative enough for making predictions on the testing dataset. Consequently, the prototypes computed on the support set $\mathcal{S}$ with Eq.~\eqref{eq: center} might not reflect the ground true. Following the transductive-learning setting \cite{liu2018learning}, we propose to incorporate the query set $\mathcal{Q}$ to refine the prototype progressively, so that one can make more accurate predictions. To this end, we propose a self-attentive prototype rectification process, where contextualized label embeddings are propagated from $\mathcal{S}$ to $\mathcal{Q}$. The overall procedure is illustrated in Figure~\ref{fig: rectification}.

\paragraph{Embedding}
We first use the feature extractor $f$ to map the input $\mathcal{S} \cup \mathcal{Q}$ to an embedding matrix $\mathbf{Z} = [\mathbf{Z}^\mathcal{S}; \mathbf{Z}^\mathcal{Q}] \in \mathbb{R}^{(NK+NM)\times d}$, where $\mathcal{S} \xrightarrow{f} \mathbf{Z}^\mathcal{S} \in \mathbb{R}^{NK \times d}$ and $\mathcal{Q} \xrightarrow{f} \mathbf{Z}^\mathcal{Q} \in \mathbb{R}^{NM \times d} $. meanwhile, the prototype matrix $\mathbf{C}=[\mathbf{c}_1; ...; \mathbf{c}_n; ...] \in \mathbb{R}^{N\times d}$ is initialized from $\mathbf{Z}^\mathcal{S}$ with Eq.~\eqref{eq: center}.
\paragraph{Prototype rectification}
Our idea is to refine the prototypes by considering the similarity information of the support set $\mathcal{S}$ and query set $\mathcal{Q}$. To achieve this, we modify the popular attention-based method~\cite{vaswani2017attention} to be suited to FSL. 

Specifically, the prototype rectification aims to aggregate information from weighted embeddings of $\mathbf{Z}$, where the attentive weight $\mathbf{A} \in \mathbb{R}^{N \times (NK+NM)}$ is given by the similarity between the current prototype $\mathbf{C}$ and the embedding $\mathbf{Z}$. As a result, the output $\mathbf{C}^*$ (the rectified prototype) is:
\begin{equation}\label{eq: oatt}
    \mathbf{C}^*=\mathbf{A} \mathbf{Z}=\textsf{Softmax}(\kappa(\mathbf{C}, \mathbf{Z}))\mathbf{Z}~.
\end{equation}
Considering the labels of $\mathcal{S}$ have been defined in advance in the local level, those labels shouldn't be ambiguous even if a new prototype is computed. So the way to compute the attention is not straightforward as shown in Eq.~\eqref{eq: oatt}, otherwise the attention matrix will be problematic as presented in \cite{ye2018learning}. Given that, we split the attention matrix $\mathbf{A}$ into two parts $\mathbf{A} = [\mathbf{A}^{\mathcal{S}},\mathbf{A}^{\mathcal{Q}}]$ with
\begin{align}
    \mathbf{A}^{\mathcal{S}}=\textsf{Softmax}(\kappa(\mathbf{C}, \mathbf{Z}^\mathcal{S}))\in \mathbb{R}^{N \times NK},\\ \nonumber \mathbf{A}^{\mathcal{Q}}=\textsf{Softmax}(\kappa(\mathbf{C}, \mathbf{Z}^\mathcal{Q}))\in \mathbb{R}^{N \times NM}.
\end{align}
where the \textsf{Softmax}$(\cdot)$ is performed on each column.
To preserve the predefined label information in the support set, we re-define the metric $\kappa(\cdot, \cdot)$ in $\mathbf{A}^{\mathcal{S}}$. Recall the label of the current data instance is $n$. The metric is defined as
\begin{align}
    \kappa(\mathbf{c}_n, \mathbf{z}_s) =
    \left\{
    \begin{array}{rll}
        +\infty &  \mbox{if} & n=y_s, \\
        -\infty & \mbox{if} & n\neq y_s.
    \end{array}
    \right.
\end{align}
where $-\infty$ and $+\infty$ indicate attention scores are 0 and 1. In this way, the attention weights in $\mathbf{A}^{\mathcal{S}}$ are non-zero only if the labels are the same. In other words, $\mathbf{A}^\mathcal{S}$ can be written as
\begin{align}
    \mathbf{A}^\mathcal{S} = [\mathbf{v}(1); ...;\mathbf{v}(n);...], \text{ for }n \in \{1, 2, ..., N\}
\end{align}
where $\mathbf{v}(n)$ is a zero-vector except for elements from $K \times (n-1)$ to $K \times n$ with value 1. With $\mathbf{A}^\mathcal{S}$ being hard-coded, the label of each support instance won't be ambiguous even if the prototype is iteratively refined. Eventually, when concatenating the hard-coded $\mathbf{A}^\mathcal{S}$ with $\mathbf{A}^\mathcal{Q}$, we re-normalize the resulting matrix per row.

To further explore the metric space and ensure the embedding is not drifted away, a residual projection layer $h: \mathbb{R}^d \xrightarrow[]{} \mathbb{R}^d$ is stacked on top of the above attention layer:
\begin{align}
    \mathbf{C} = h(\mathbf{C}^*) + \mathbf{C}, \mathbf{Z} = h(\mathbf{Z}) + \mathbf{Z}
\end{align}

\begin{table*}[t!]
    \centering
    \scalebox{0.95}{
    \begin{tabular}{l|c|c|c|c|c|c}
    \toprule
    \multirow{2}{*}{{\bf Models}} & \multirow{2}{*}{{\bf Backbone}} & \multirow{2}{*}{{\bf FT}} & \multicolumn{2}{c|}{\textbf{\emph{miniImageNet}}} & \multicolumn{2}{c}{\textbf{\emph{tieredImageNet}}} \\
        &          &        &  1-shot    &    5-shot      &   1-shot        &     5-shot     \\ \midrule
    Prototypical Net~\cite{snell2017prototypical}    & ConvNet  &    \xmark   & 49.42\tiny{$\pm0.78$} &  68.20\tiny{$\pm0.66$}  & 53.31\tiny{$\pm0.89$} & 72.69\tiny{$\pm0.74$}         \\ 
    Relation Net~\cite{sung2018learning}    & ConvNet  &    \xmark   & 50.44\tiny{$\pm0.82$} & 65.32\tiny{$\pm0.70$} &    54.48\tiny{$\pm0.93$}   & 71.32\tiny{$\pm0.78$} \\
    TPN~\cite{liu2018learning}    & ConvNet  &    \xmark   & 55.51\tiny{$\pm0.86$} & 69.86\tiny{$\pm0.65$} & 59.91\tiny{$\pm0.94$} & 73.30\tiny{$\pm0.75$} \\ 
    FEAT~\cite{ye2018learning}    & ConvNet  &    \cmark   & 55.75  & 72.17   &  --   &   --       \\ 
    EPNet~\cite{rodriguez2020embedding}    & ConvNet  &    \cmark   & 59.32\tiny{$\pm0.84$} &  72.95\tiny{$\pm0.63$}  & 60.70\tiny{$\pm0.97$} & 73.91\tiny{$\pm0.74$}         \\ 
    DFMN-MCT~\cite{kye2020transductive}    & ConvNet  &    \xmark   & 64.65\tiny{$\pm0.89$} & 75.96\tiny{$\pm0.54$} & 65.66\tiny{$\pm0.98$} & 75.72\tiny{$\pm0.61$} \\ \midrule
    \rowcolor{Gray}
    ReMP       & ConvNet  &  \xmark  & \textcolor{tblue}{\bf 66.21\tiny{$\pm0.36$}}  & \textcolor{tblue}{\bf 76.50\tiny{$\pm0.29$}} & \textcolor{tblue}{\bf 67.12\tiny{$\pm0.87$}} &  \textcolor{tblue}{\bf 76.43\tiny{$\pm0.50$}}  \\
    \midrule
    \midrule
    TADAM~\cite{oreshkin2018tadam}    & ResNet  &    \xmark   & 58.50\tiny{$\pm0.30$} & 76.70\tiny{$\pm0.30$} & -- & -- \\ 
    TPN~\cite{liu2018learning}    & ResNet  &    \xmark   & 59.46 & 75.65 & -- & -- \\ 
    FEAT~\cite{ye2018learning}    & ResNet  &    \cmark   & 62.60  & 78.06   &  --   &   --       \\
    CAN~\cite{hou2019cross}    & ResNet  &  \xmark   & 67.19\tiny{$\pm0.55$} &  80.64\tiny{$\pm0.35$}  & 73.21\tiny{$\pm0.58$} & 84.93\tiny{$\pm0.38$}         \\ 
    EPNet~\cite{rodriguez2020embedding}    & ResNet  &    \cmark   & 70.74\tiny{$\pm0.86$} &  81.52\tiny{$\pm0.58$}  & 78.50\tiny{$\pm0.86$} & 87.48\tiny{$\pm0.69$}         \\ 
    DFMN-MCT~\cite{kye2020transductive}    & ResNet  &    \xmark   & 78.30\tiny{$\pm0.81$} & 86.48\tiny{$\pm0.42$} & 80.89\tiny{$\pm0.84$} & 87.30\tiny{$\pm0.49$} \\ \midrule
    \rowcolor{Gray}
    ReMP        & ResNet  &    \xmark   & \textcolor{tblue}{\bf 79.25\tiny{$\pm0.31$}} & \textcolor{tblue}{\bf 87.01\tiny{$\pm0.29$}} &  \textcolor{tblue}{\bf 82.01\tiny{$\pm0.71$}}& \textcolor{tblue}{\bf87.92\tiny{$\pm 0.38$}} \\
    \bottomrule
    \end{tabular}
    }
    \vspace{2mm}
    \caption{Comparison with state-of-the-art methods on 5-way classification (FT: fine-tuning).}
    \label{tab: main_results}
\end{table*}
\paragraph{Repulsive attention}
In an episode of the standard $N$-way FSL classification scenario, each of the $N$ classes is associated with $M$ queries. With the attention mechanism defined above, the prototype rectification for a specific class might be affected by the other $N(M-1)$ queries beyond this class. Intuitively, queries too far away from the current instance is not expected to interact with the current instance. As a result, we propose a repulsive self-attention mechanism to refine the attention scores. Specifically, given a threshold $\beta$, we refine the attention score as:
\begin{align}\label{eq: mask}
    \mathbf{A}[\mathsf{mask}]  = -\text{min}(\mathbf{A}), \text{ where } \mathsf{mask} = (\mathbf{A} < \beta).
\end{align}
Also, a larger $\beta$ tends to enforce weaker repulsive force. The enforced negative attention score, which corresponds to some dissimilar queries, can induce a repulsive force that prevents a prototype from being pulled away by these queries, as demonstrated in Figure~\ref{fig: rectification}.

In our experiment, we stack $L$ layers of the rectification process described above, corresponding to an $L$-layer attention mechanism, which has been demonstrated effective in natural language processing (NLP) tasks~\cite{vaswani2017attention,lan2019albert}.

\section{Experiments}
In this section, we evaluate the proposed ReMP to seek answers for the following questions:
\textbf{\texttt{Q1}}: How does ReMP perform on standard benchmarks compared to SoTA?
\textbf{\texttt{Q2}}: Does ReMP indeed reduce the inconsistency between training and testing?
\textbf{\texttt{Q3}}: How much does each component of ReMP contribute to the performance? 

\subsection{Experimental Settings}
\paragraph{Datasets} In the main part, we consider two standard FSL datasets, \textit{miniImageNet}, \textit{tieredImageNet} and \textit{CIFAR-FS}.
\begin{itemize}
    \item \textit{miniImageNet} is a subset of ILSVRC-12~\cite{deng2009imagenet} proposed in~\cite{vinyals2016matching} for FSL evaluation. It consists of 100 classes with 600 images per class. We follow the standard protocol to divide the dataset into three subsets:  $\mathcal{D}^{\text{train}}$ of 64 classes,  $\mathcal{D}^{\text{test}}$ of 20 classes and $\mathcal{D}^{\text{val}}$ of 16 classes. 
    \item \textit{tieredImageNet} is an alternative subset of ILSVRC-12 prepared for more challenging FSL evaluation. It has a hierarchical structure of 34 coarse categories with fine-grained classes. There are three subsets: $\mathcal{D}^{\text{train}}$ with 20 categories and 351 classes, $\mathcal{D}^{\text{test}}$ with 8 categories and 160 classes, and $\mathcal{D}^{\text{val}}$ with 6 categories and 97 classes.
    \item \textit{CIFAR-FS} is based on CIFAR-100~\cite{krizhevsky2009learning}. It is split into 64 training classes, 16 validation classes and 20 testing classes.
\end{itemize}

\paragraph{Implementation details}
All experiments are conducted in PyTorch-1.3 on NVIDIA TITAN XP (12GB) platform. 
Following~\cite{kye2020transductive,sung2018learning,lee2019meta}, we consider two standard embedding modules ($f$) as backbones: a 4-layer ConvNet with 64 channels per layer and a 12-layer ResNet, as well as standard data augmentation (including random resized crop and horizontal flip). Following~\cite{sung2018learning,kye2020transductive}, in each episode, the number of queries in each class is set to $M=15$. Cosine similarity is used in~\eqref{eq: ll_global} and negative Euclidean distance is used in~\eqref{eq: ll_local}. All models are trained with a Stochastic Gradient Descent (SGD) optimizer with momentum 0.9. The initial learning rate is set to $0.1$ and the weight decay is $5e^{-3}$. We decay the learning rate by a factor of 10 every 25000 iterations until convergence. We stack $L=2$ layers in training and $L=10$ layers in testing. The default loss balance factor $\alpha=0.1$. All reported accuracy results are averaged over 600 test episodes with 95\% confidence intervals. 
\paragraph{Baseline Methods} We compare our approach with seven baselines. All baselines, except for Prototypical Net~\cite{snell2017prototypical}, Relation Net~\cite{sung2018learning} and TADAM~\cite{oreshkin2018tadam}, are transductive-learning based approaches. Specifically, FEAT~\cite{ye2018learning} and CAN~\cite{hou2019cross} use embedding propagation while TPN~\cite{liu2018learning}, EPNet~\cite{rodriguez2020embedding} and DFMN-MCT~\cite{kye2020transductive} are based on label propagation. We also note that, $(\RN{1})$ TADAM introduces a co-training method that takes the parametric linear classifier as an auxiliary task whose importance is decayed, however, it is still decoupled from the primary FSL task; $(\RN{2})$ FEAT borrows the entire \textsf{Transformer} architecture from~\cite{vaswani2017attention}. This tends to break both the pretrained embedding space and ignore the predefined label information (if used in a transductive manner), resulting in a limited performance gain; $(\RN{3})$ CAN~\cite{hou2019cross} also combines the local and global losses, both of which are defined with cosine similarity. However, we demonstrate that it is not necessary to use the same metric as the global loss and the local loss don't necessarily share the same ``prototypes''.

\subsection{Comparison with SoTA methods}
\begin{figure}[!htbp]
    \centering
    \begin{tabular}{c}
        \includegraphics[width=0.65\linewidth]{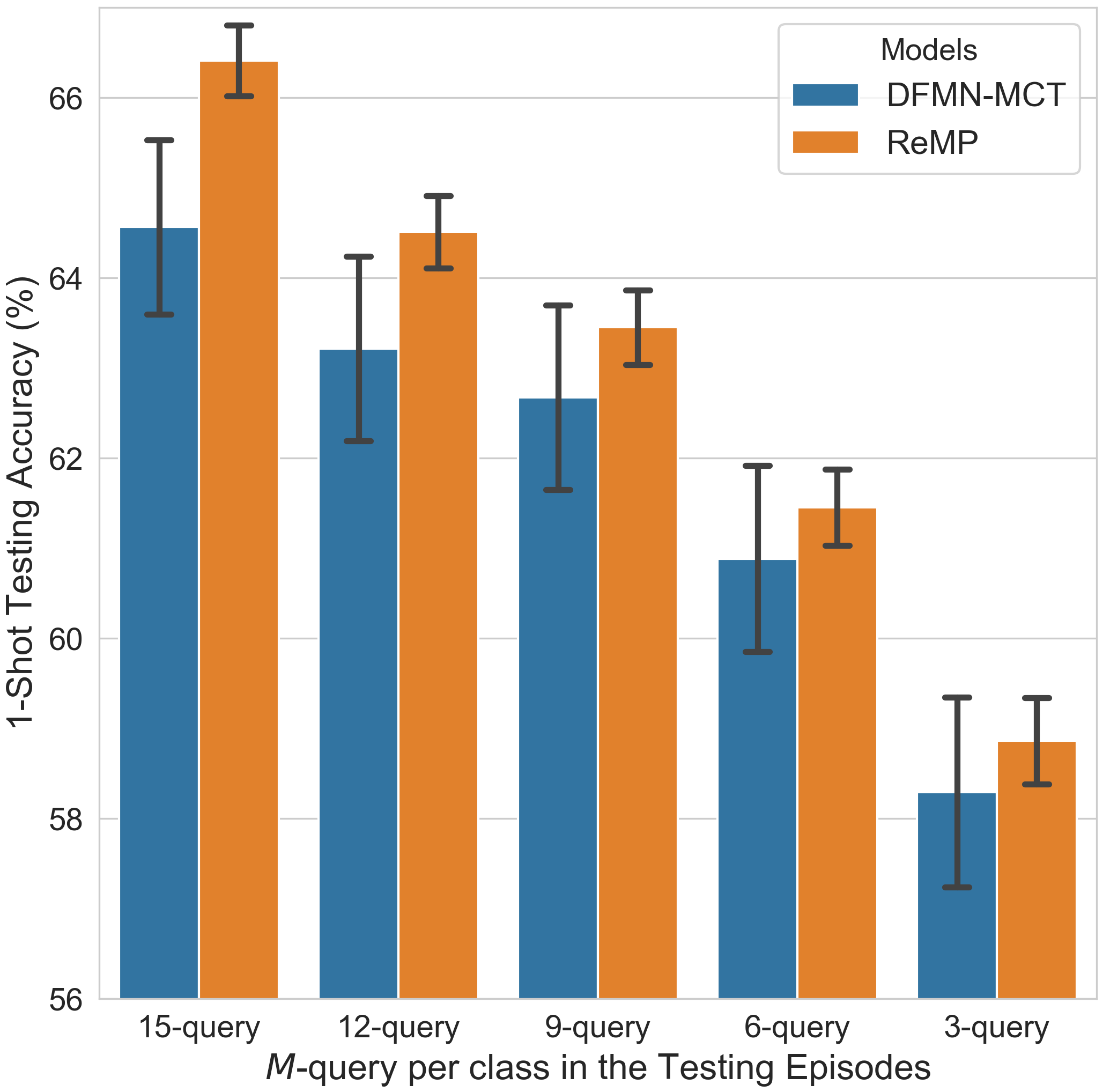} \\ 
        \vspace{2mm}
        \includegraphics[width=0.65\linewidth]{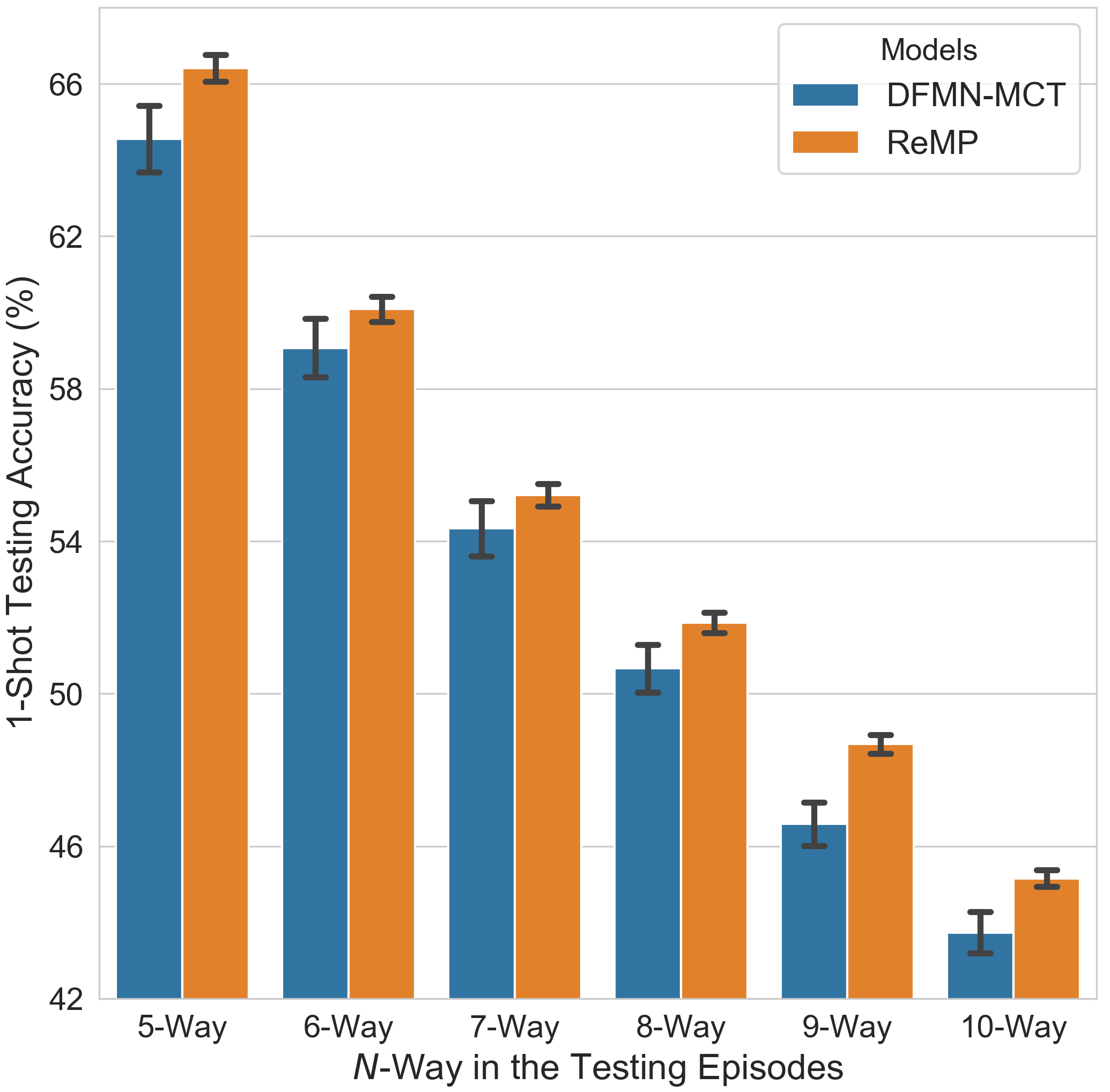}
    \end{tabular}
    \vspace{2mm}
    \caption{Comparison showing the effect of: ({\bf Top}) the number of test queries per class and ({\bf Bottom}) the number of classes per testing episode for 1-shot protocol on the {\it miniImageNet}.}
    \label{fig:1-shot-n-way}
\end{figure}

The main results are shown in Table~\ref{tab: main_results}. As is seen, ReMP obtains the highest FSL classification accuracy as well as the lowest deviations in most settings by a large margin. Table~~\ref{tab: cifar} shows the 1-shot and 5-shot accuracy results on \textit{CIFAR-FS}.
\begin{table}[!htbp]
    \centering
    \scalebox{0.85}{
    \begin{tabular}{l|c|c|c}
    \toprule
        \textbf{Models} & \textbf{Backbone} & \textbf{1-shot} & \textbf{5-shot} \\ \midrule
        ProtoNet & ConvNet & 55.50\tiny{$\pm0.70$} & 72.00 \tiny{$\pm0.60$} \\ 
        MetaOpt-SVM & ResNet & 72.00\tiny{$\pm0.70$} & 84.20\tiny{$\pm0.50$} \\ 
        DFMN-MCT & Resnet & 87.51\tiny{$\pm0.48$} & 90.23\tiny{$\pm0.63$} \\ \midrule
        \rowcolor{Gray}
        ReMP & ResNet & \textcolor{tblue}{\bf87.83\tiny{$\pm0.31$}} & 90.17\tiny{$\pm0.50$} \\ \bottomrule
    \end{tabular}
    }
    \vspace{2mm}
    \caption{Comparison with various baselines on CIFAR-FS.}
    \label{tab: cifar}
\end{table}
\paragraph{{\it N}-way {\it M}-query learning}
We conduct further analysis to determine the effect of the number of queries per class and the number of classes per testing episodes. Decreasing the number of queries and increasing the number of classes could both affect the model's prediction accuracy. The related work DFMN-MCT, which holds the current state-of-the-arts, is chosen as a strong baseline. ReMP distinguishes from DFMN-MCT in a rectified metric objective and an efficient repulsive attention-based propagation layer. As seen in Figure~\ref{fig:1-shot-n-way}, ReMP consistently outperforms DFMN-MCT under two sets of scenarios, endowing both higher accuracy and lower deviation.

\subsection{Why does ReMP improve performance?}
\paragraph{On the rectification of learning objective}
We present four scenarios covering the impact of training schedules and metric options for the demonstration: $(\RN{1})$ Cooperative training - apply $\mathcal{L}_{full}$~\eqref{eq: full}; $(\RN{2})$ Pretrain and finetune - apply $\mathcal{L}_{global}$~\eqref{eq: global} then use $\mathcal{L}_{local}$~\eqref{eq: local}; $(\RN{3})$ Local matching - only apply $\mathcal{L}_{local}$; and $(\RN{4})$ Global matching - only apply $\mathcal{L}_{global}$. The main results are shown in Tab.~\ref{tab: metric}, from which we can conclude:
\begin{itemize}
    \item Global matching stimulates a more discrminative metric space, as it is observed that the accuracy in (b) is always higher than the corresponding one in (c), regardless of the metrics.
    \item Local matching mitigates the discrepancy between training and testing as seen in (a) and (d). The reason lies in the improved transferability from training to testing, as they are being optimized towards exactly the same goal in \eqref{eq: ll_local}.
\end{itemize}
\begin{figure}[!htbp]
    \centering
    \includegraphics[width=0.6\linewidth]{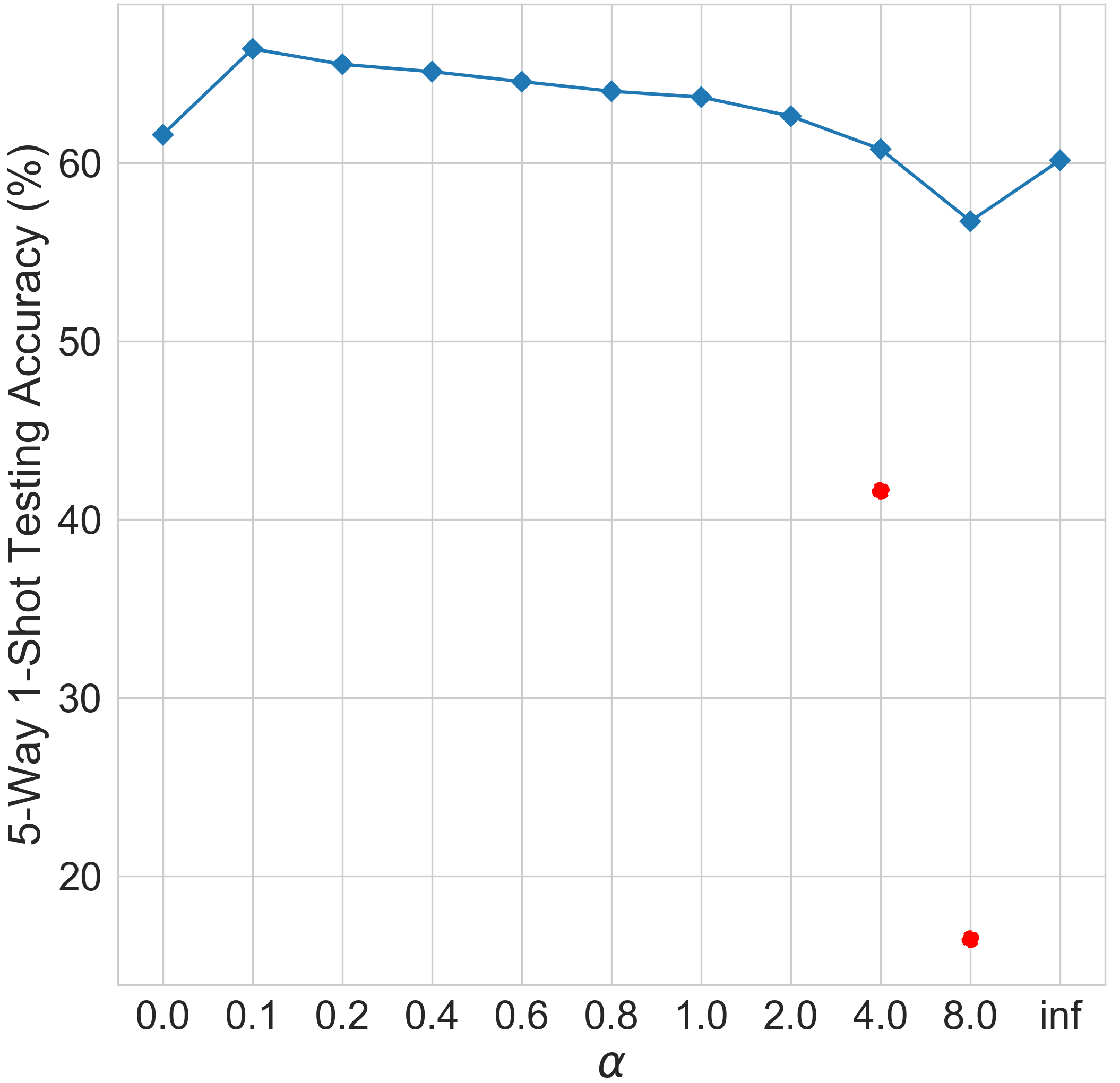}
    \vspace{2mm}
    \caption{The impact of $\alpha$ under Euclidean distance.}
    \label{fig: alpha}
\end{figure}

In addition, comparing (a) with (b) and (d), \textbf{we find that cooperative learning surprisingly obtains the best performance when Cosine similarity and Euclidean distance are applied at the same time}, which nearly closes the gap between the two measures. We hypothesize that the metric-based FSL problem is fundamentally difficult without inductive biases on both the models and the data. The most suitable metric space can not be easily identified. Figure~\ref{fig: t-sne} in Appendix~\ref{sec: appendix_tsne} shows the t-SNE~\cite{maaten2008visualizing} visualization of embedding spaces. The observation is consistent with Tab.~\ref{tab: metric}. Above findings provide strong support for Remark \ref{remark} to answer \textbf{\texttt{Q2}}.
\begin{table}[!htbp]
    \centering
    \begin{tabular}{c}
         \includegraphics[width=0.6\linewidth]{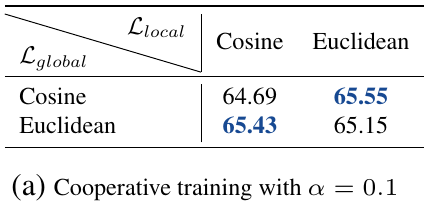}  \\
         \includegraphics[width=0.6\linewidth]{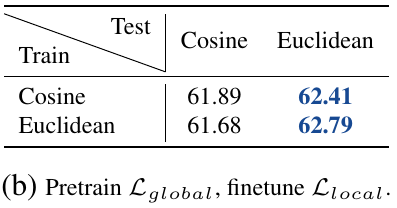} \\
         \includegraphics[width=0.6\linewidth]{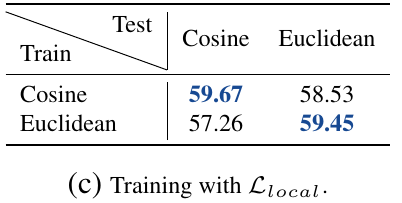} \\
         \includegraphics[width=0.6\linewidth]{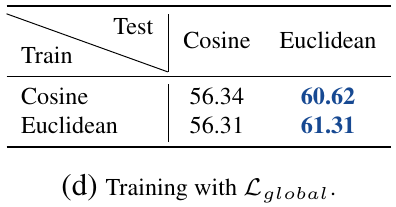} 
    \end{tabular}
    \vspace{1mm}
    \caption{The impact of similarity $\kappa(\cdot, \cdot)$ and training schedule. Results are averaged over 10 independent runs of 6000 episodes.}
    \label{tab: metric}
\end{table}
We also visualize the impact of the balance factor $\alpha$ in Figure~\ref{fig: alpha}, where two red dots appeared as the unstable model training tries (about 6 out of 10). When $\alpha=0$, global matching itself will cause inconsistency issues. As $\alpha$ gets a bit larger, global and local prototypes try to reach a consensus to induce an optimal model. If $\alpha$ is very large, instability issues will be triggered where global and local prototypes struggle to fit each other. When $\alpha \xrightarrow[]{} \inf$, the local matching itself won't find a discriminative space as the global matching does. 

\begin{figure}[!h]
    \centering
    \includegraphics[width=\linewidth]{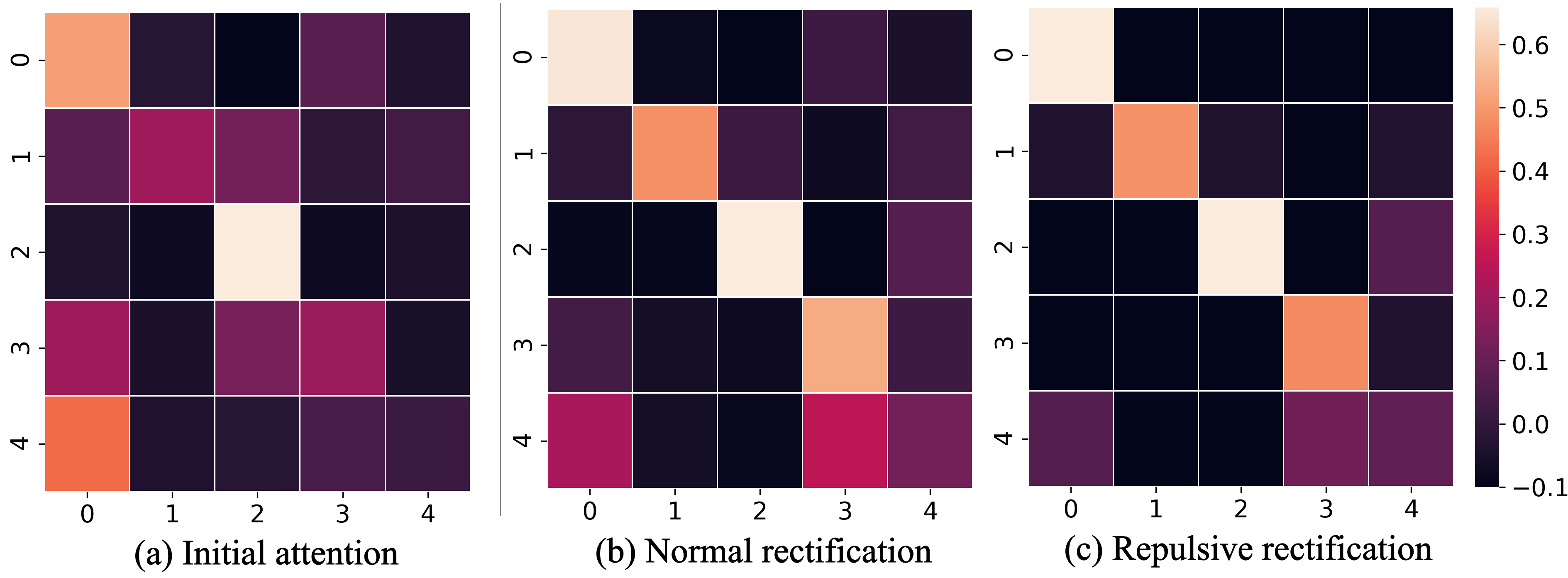}
    \caption{Visualization of the rectification process for 5-way 1-shot classification on \textit{miniImageNet}. In a heatmap, the $x$-axis and $y$-axis represent class types for prototypes and queries, respectively. For example, the value in (row 1, column 2) represents the similarity between query of class 1 and prototype of class 2. For better classification, the diagonal values should be the highest in each row.}
    \label{fig: focal_move}
\end{figure}

\paragraph{Repulsive attention}
We empirically define a layer-wise threshold $\beta_l=1.5/N(L-l)$ throughout our experiments, where $l$ is the layer number of a stacked propagation. $\beta_l$ is expected to gradually increase as we rectify the prototype with more layers. We visualize the rectification process using attention heatmap in Figure~\ref{fig: focal_move}. Apparently, the iterative rectification process leads to a better decision boundary. And generally, the boundary can be further optimized by the repulsive rectification as seen in Figure~\ref{fig: rectification}(c). 

\paragraph{Ablate the component}\label{sec: appendix_ab}
To anwer \textbf{\texttt{Q3}}, we conduct an ablation study to examine the contribution of each component to our model. As it can seen in Table~\ref{tab: ablation}, the transductive FSL obviously outperforms the inductive implementation. More importantly, the global matching indeed encourages a more discriminative metric space than the local matching does. In summary, the cooperative learning can give the best performance. 
\begin{table}[!htbp]
    \centering
    \scalebox{0.8}{
    \begin{tabular}{l|l}
    \toprule
        \textbf{Ablation} & \textbf{Accuracy} \\ \midrule
        ReMP &  66.21\tiny{$\pm0.36$} \\ \midrule
        ReMP w/o repulsive attention &  65.55\tiny{$\pm0.36$} \\
        ReMP w/o $\mathcal{L}_{local}$ &  60.62\tiny{$\pm0.37$} \\
        ReMP w/o $\mathcal{L}_{global}$ &  58.53\tiny{$\pm0.36$} \\
        ReMP w/o contextualization (inductive) & 56.13\tiny{$\pm0.78$} \\
    \bottomrule
    \end{tabular}
    }
    \vspace{2mm}
    \caption{Ablation study results with 95\% confidence intervals on 1-shot 5-way classification on \textit{miniImageNet} (backbone: ConvNet-64).}
    \label{tab: ablation}
\end{table}
\section{Conclusions}
In this paper, we have presented a rectified metric propagation (ReMP) framework for few-shot learning. To cope with the inconsistency issues between training and testing in metric-based FSL, we propose a label-aligned learning objective to close the gap, where a discriminative and transferable embedding space is induced. To further enhance the quality of the prototype in the testing phase, we propose a contextualized label embedding module, which progressively propagates labels to the query set basis in a repulsive self-attention mechanism. Empirically, ReMP achieves state-of-the-art results on various FSL datasets. 

\clearpage
{\small
\bibliographystyle{ieee_fullname}
\bibliography{egbib}
}

\clearpage
\appendix\label{sec: appdendix}

\section{Box-Plot Comparison}
\begin{figure}[!htbp]
    \centering
    \scalebox{0.9}{
    \begin{tabular}{c}
        \includegraphics[width=0.98\linewidth]{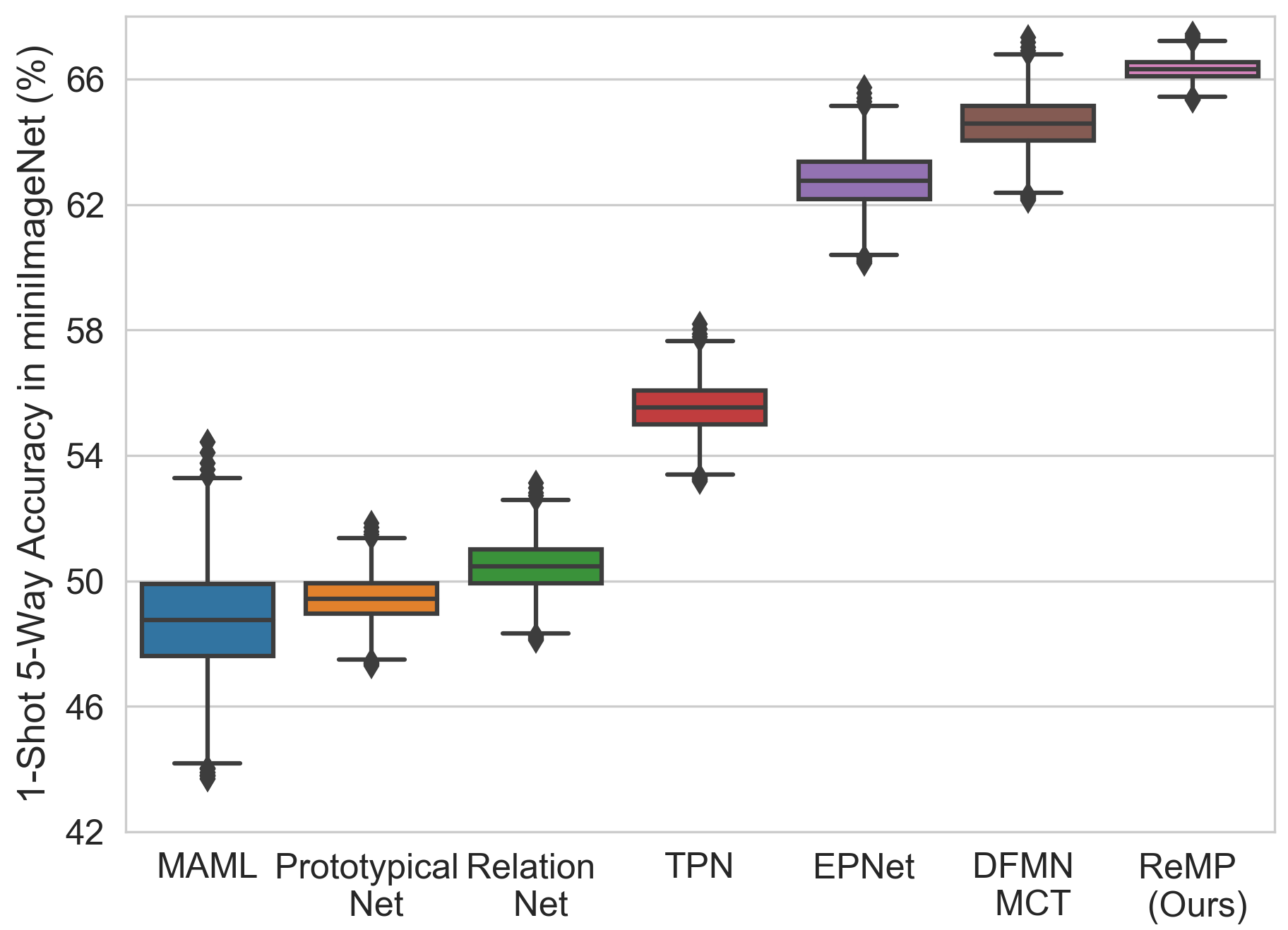} \\
        \includegraphics[width=0.98\linewidth]{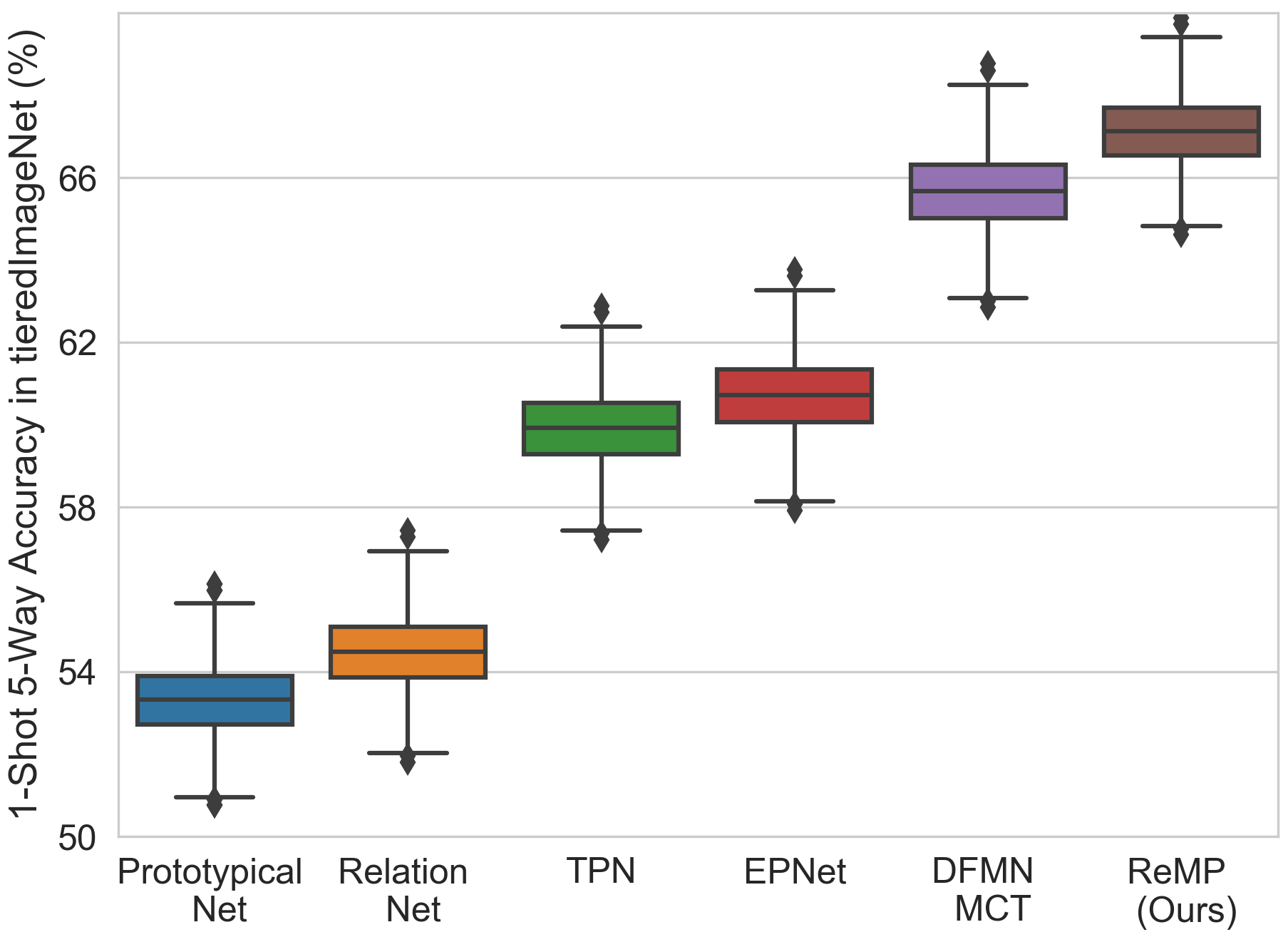}
    \end{tabular}
    }
    \caption{Performance illustration of popular few-shot learning models~\cite{finn2017model,snell2017prototypical,sung2018learning,liu2018learning,rodriguez2020embedding,kye2020transductive} in 1-shot 5-way (Left) miniImageNet~\cite{vinyals2016matching} and (Right) tieredImagenet classification. To be fair, all models are built on the backbone 4-layer ConvNet-64 except the MAML, which uses a 4-layer ConvNet-48. The boxes indicate the interquartile range of the accuracy while the notches show the median and its 95\% confidence interval. Whiskers denote the $1.5\times$ interquartile range which captures 99.3\% of the probability mass for a normal distribution~\cite{dhillon2019baseline}.}
    
    \label{fig:accuracy_cmp}
\end{figure}

We use a box-plot to visualize the improvement our proposed ReMP has made over the previous SoTA methods in Figure~\ref{fig:accuracy_cmp}. As is seen, our approach can achieve a higher accuracy while attaining a lower standard deviation.

\section{Visualization of Metric Space}\label{sec: appendix_tsne}
We visualize the metric space with t-SNE in Figure~\ref{fig: t-sne}. Cosine similarity is applied both for training and testing in this part. The class separability follows the order: Coopertaive training $>$ Pretrain-Finetune $>$ Local matching $>$ Global matching, which is consistent with the results in Table~\ref{tab: metric}.
\begin{table}[!htbp]
    \centering
    \begin{tabular}{c}
         \includegraphics[width=0.6\linewidth]{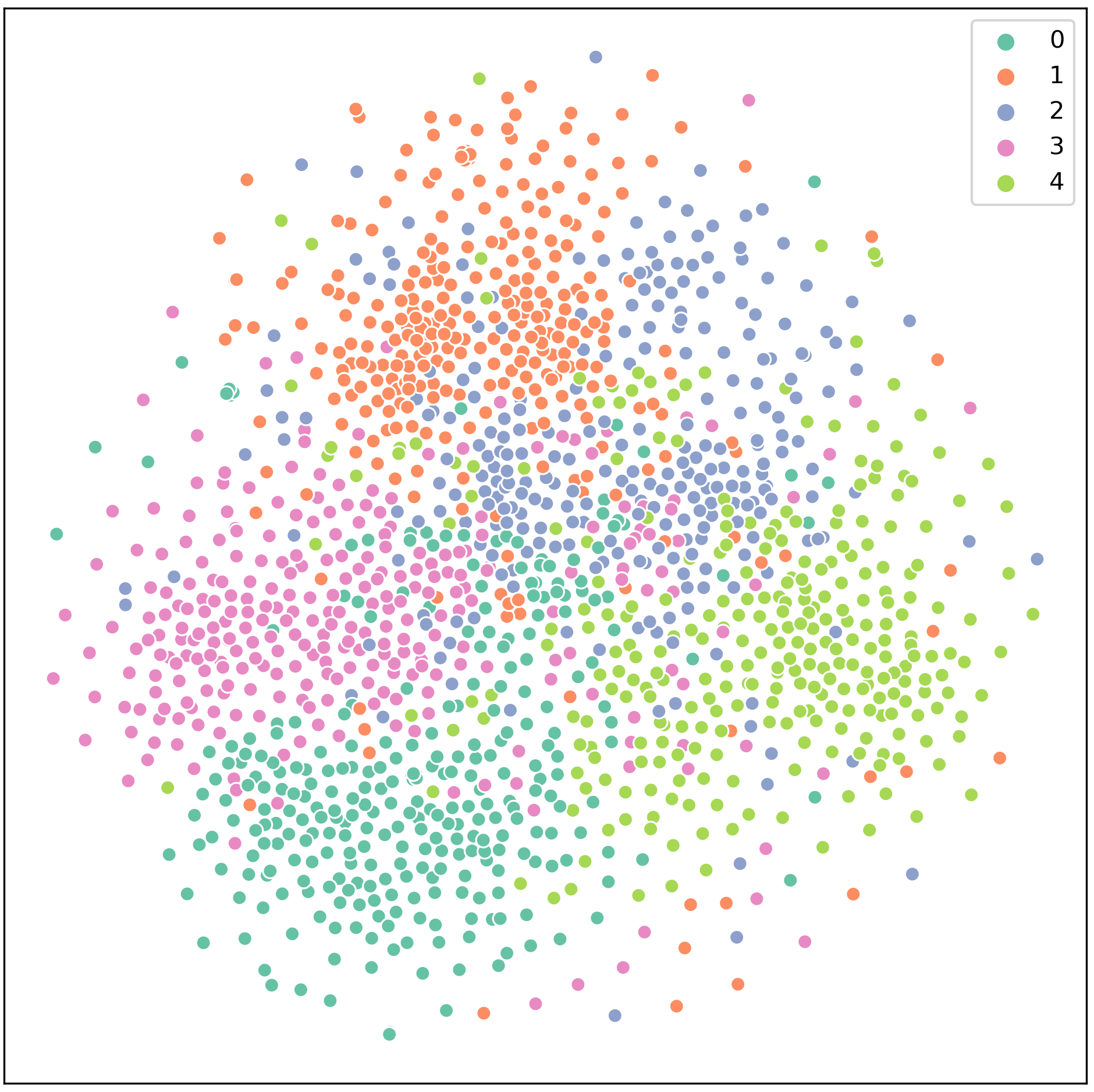}  \\
         \includegraphics[width=0.6\linewidth]{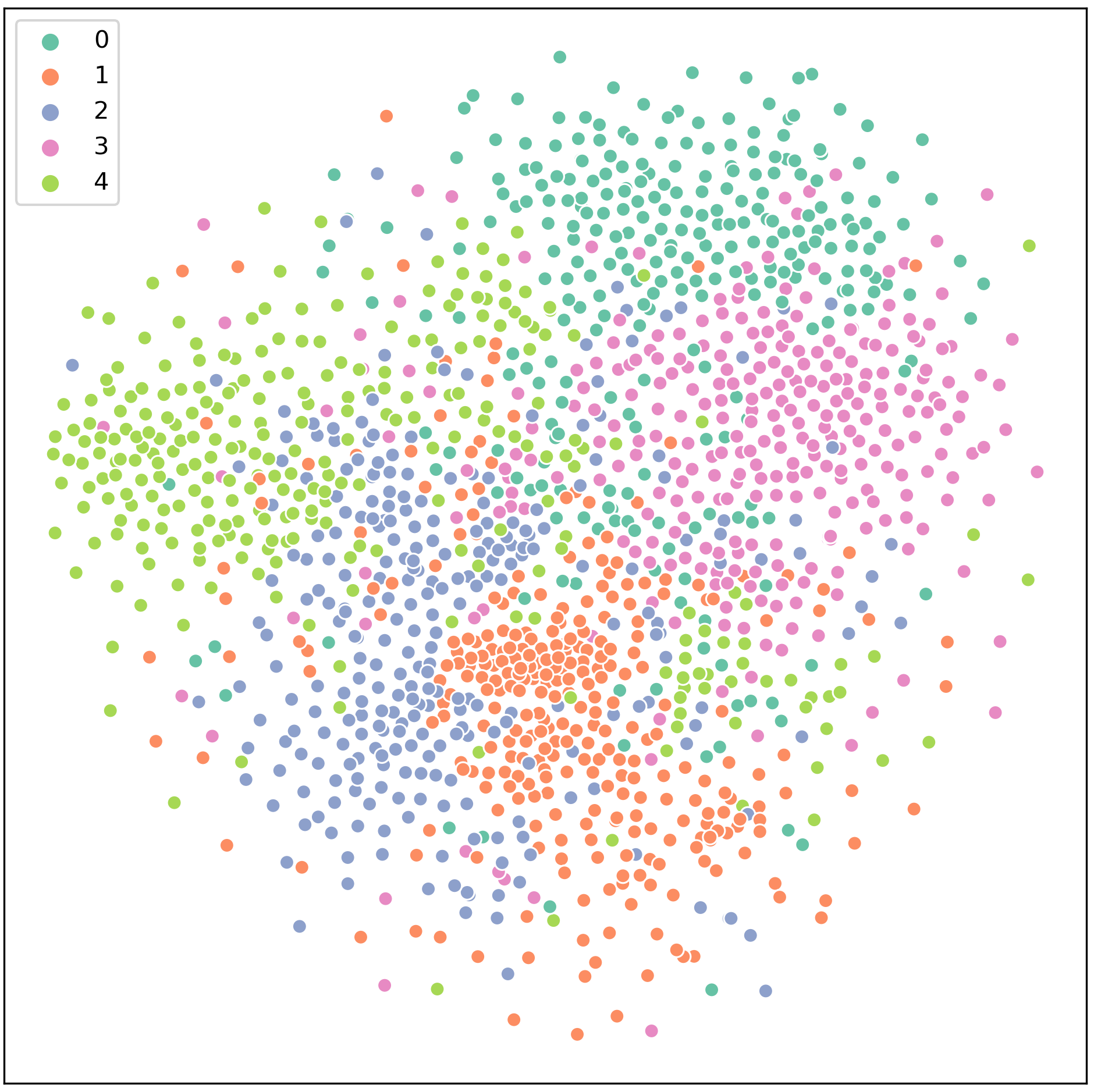} \\ 
         \includegraphics[width=0.6\linewidth]{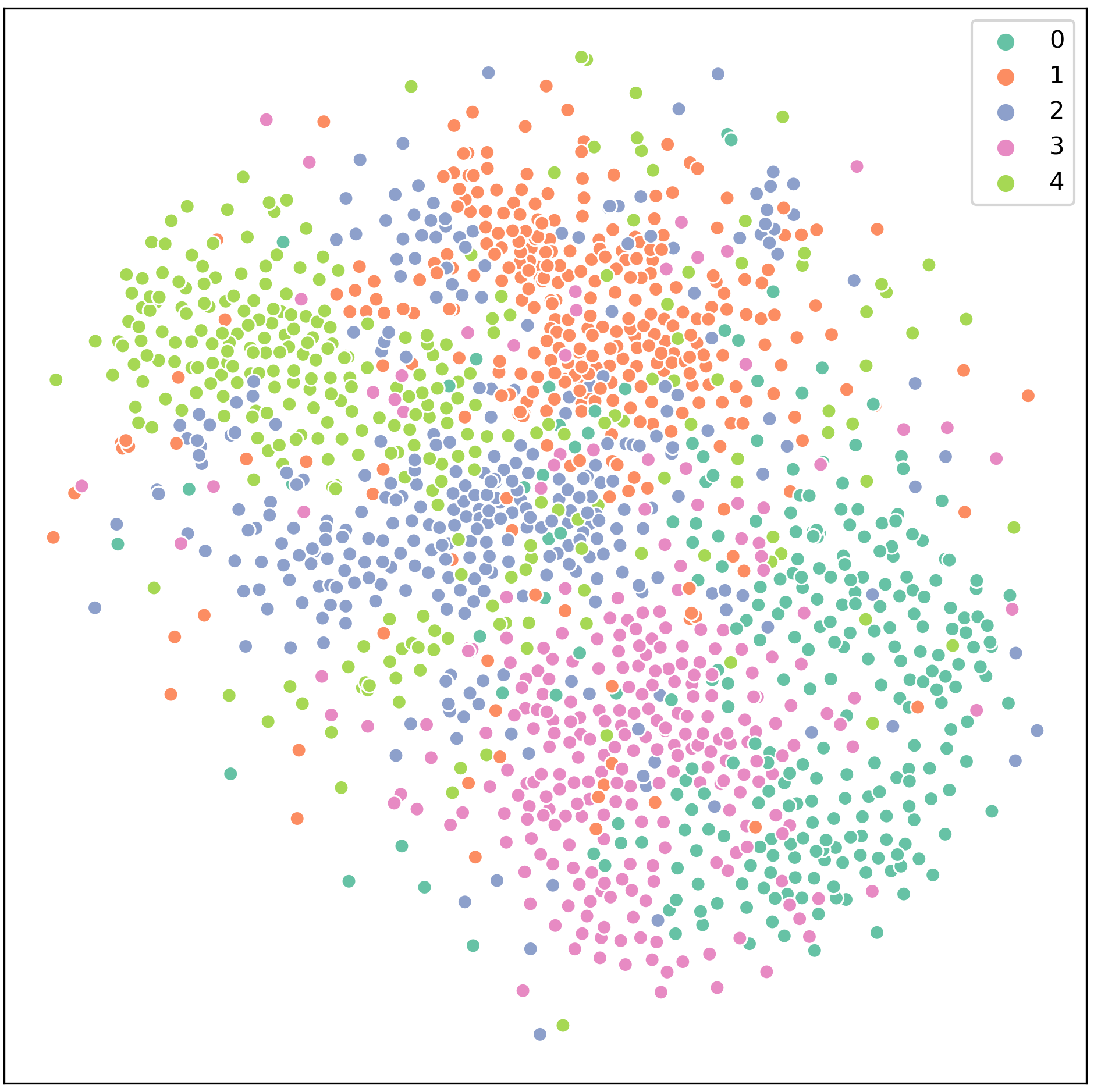} \\
         \includegraphics[width=0.6\linewidth]{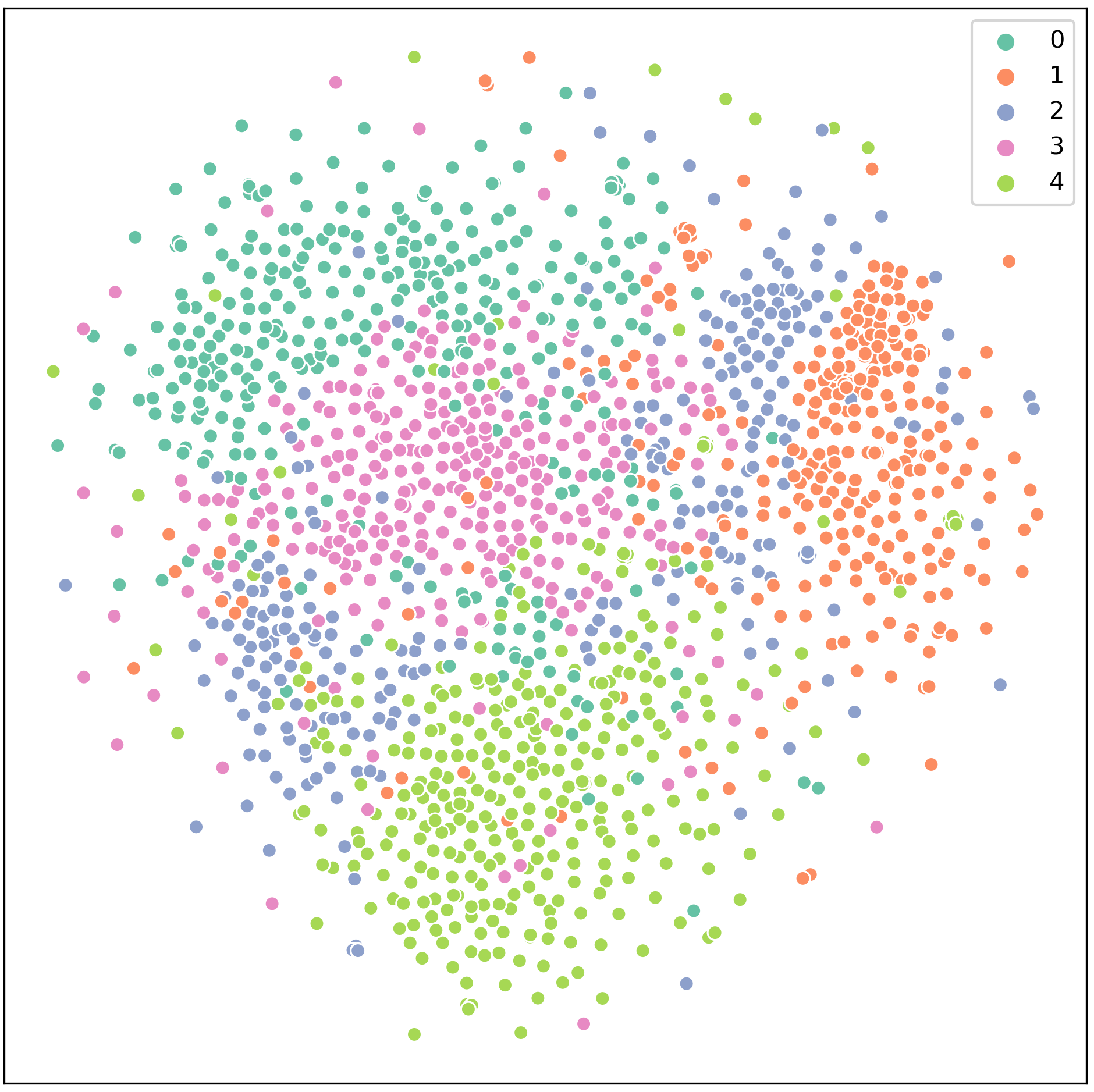}
    \end{tabular}
    \caption{t-SNE visualizations of metric space for 5-way 1-shot protocol based on ConvNet-64 for \textit{miniImageNet} classification. The first 5 classes are selected from $\mathcal{D}^{\text{test}}$. From top to bottom: Cooperative training, Pretrain then finetune, Pretrain then finetune and Global matching.}
    \label{fig: t-sne}
\end{table}


\end{document}